# LLM Parkinsonism: Executive-Control Failure, Token-Inefficient Persistence, and an Uncertainty-Aware Global Executive Control Architecture for Autonomous Language-Model Agents

Dongsheng Xiao[1]*, Zeyuan Wang[2], Xuzhe Xia[3], Bo Zhao[4], Yankai Cao[5]

[1] Queensland Brain Institute, The University of Queensland, Brisbane, Queensland, Australia

[2] School of Cyber Security, Northwestern Polytechnical University, Xi'an, Shaanxi, China

[3] Department of Electrical and Computer Engineering, The University of British Columbia, Vancouver, British Columbia, Canada

[4] Neurointelligence Labs, Vancouver, British Columbia, Canada

[5] Department of Chemical and Biological Engineering, The University of British Columbia, Vancouver, British Columbia, Canada

dongsheng.xiao@uq.edu.au, wangzeyuan55@gmail.com, xiaxuzhe2019@gmail.com, zhaobo.cs@gmail.com, yankai.cao@ubc.ca

*Corresponding author

**Abstract**

Large language models (LLMs) can plan, use tools, write code, and execute long-horizon workflows, yet strong local competence does not guarantee project-level executive control. Agents may continue acting after the original objective is satisfied, producing low-value refinements, repeated verification, and repairs to self-created complexity. We use LLM Parkinsonism as a narrowly defined, non-clinical metaphor for this pattern of persistent action despite diminishing task-level value. We argue that the problem is not explained by autoregressive next-token prediction alone, but more directly by concentrating proposal generation, scope interpretation, progress assessment, and stopping authority within the same self-conditioned loop. We therefore introduce Global Executive Control (GEC) v0.2, an uncertainty-aware governance architecture that separates action generation from project-level control. In a 24,000-episode matched-candidate benchmark under a common 40,000-token ceiling, a first-candidate baseline achieved 67.42% hard-goal success, a candidate-set local control achieved 96.53%, and GEC achieved 96.57%. The candidate-set control shows that access to multiple candidate actions explains most of the success gain; relative to that control, GEC preserved success while reducing mean token use from 19,782 to 12,574 (36.4%) and restricted mean tokens to completion at the 40,000-token ceiling from 16,136 to 13,114 (18.7%), while eliminating measured pre-completion drift and sharply reducing gross complexity. Governance-overhead sensitivity remained favorable through an additional 500 synthetic governance tokens per cycle. These mechanistic simulations support explicit governance of scope, evidence, resource use, and stopping, while live-model validation remains necessary.

## 1 Introduction

The dominant narrative of contemporary LLM progress is capability expansion. Models have become increasingly competent at coding, database design, tool use, web navigation, planning, debugging, analysis, and multimodal interaction. Agentic systems amplify these capabilities by embedding LLMs in iterative loops that observe, reason, act, observe again, and continue until a task is judged complete. ReAct established an influential pattern for interleaving reasoning and action [1], while Reflexion demonstrated how language agents can use feedback and verbal reflection to modify subsequent behavior [2]. Instruction-following optimization further encourages models to respond usefully to user requests [3].

A qualitatively different question emerges once an agent can do many things: who decides which of those things should still be done?

A model may competently implement architecture, databases, APIs, front ends, deployments, monitoring, and disaster recovery while entering an increasingly narrow tail of activity. A core system can be functionally complete, yet the agent continues to add optional layers, discover edge cases in those layers, write tests for the edge cases, and repair failures in abstractions that it created. The local engineering can remain coherent while the global decision to continue becomes poor.

A motivating example is a two-node disaster-recovery watchdog. The initial Goal Contract can be stated compactly:

- if node A becomes unavailable, emit A NODE DOWN ALERT;
- if node B becomes unavailable, emit B NODE DOWN ALERT;
- if both are unavailable, both node-level conditions remain observable; and
- on recovery, restore the correct healthy state.

An agent may nevertheless introduce site-level alert aggregation: when A and B are simultaneously unavailable, suppress both individual alerts and emit a single SITE DOWN alert. That apparently sophisticated improvement creates correlation state, timing windows, suppression rules, synthetic tests, new failure modes, and deployment gates. The agent can then spend substantial effort debugging a subsystem that the original goal never required.

This motivates a distinction between problem-solving intelligence and executive-control intelligence. The former asks how to perform the next operation. The latter asks whether the operation remains part of the governed objective, whether an enabling step is causally justified, whether complexity is worth its cost, whether evidence is still valid after later changes, and whether stopping is now rational.

The distinction matters economically. Agent quality cannot be characterized only by eventual success when inference and tool use are resource-metered. Cost-aware evaluation has emerged as an important concern [4-8]. Long-horizon benchmarks expose difficulties with constrained global planning [9]; overthinking work shows diminishing or negative marginal returns from additional reasoning [10,11]; workflow scheduling can reduce unnecessary computation [12]; infinite agentic loops can create runaway model and tool calls [13]; and externally grounded verification work shows that self-evaluation can mistake activity for progress [14].

This paper makes six contributions. First, we define LLM Parkinsonism as a non-clinical trajectory-level construct rather than a synonym for verbosity. Second, we propose a systems hypothesis centered on collapsed executive authority. Third, we replace an overly rigid criterion-closing scope rule with a governed causal-link model that admits prerequisites, verification, and risk mitigation. Fourth, we introduce GEC v0.2 with state-versioned evidence, independent scope adjudication, governed Goal Contract amendments, and state-level economic stopping. Fifth, we revise the metrics to separate pre-completion drift from post-completion persistence and replace conditional completion time as a primary comparison with a budget-restricted completion-cost endpoint. Sixth, we release LPB v0.2 with matched candidate sets, a candidate-set local control, budget and ablation studies, governance-overhead sensitivity, noise robustness, and beneficial-soft-work probes.

The central thesis is simple: A model that can always propose another reasonable action is not necessarily a model that knows whether another action is worth taking.

## 2 Related Work

### 2.1 Agent loops and self-improvement

ReAct and Reflexion exemplify iterative agent designs in which model outputs alter subsequent state and context [1,2]. Such loops are powerful because they support adaptation, but continuation creates a control problem: local plausibility of the next action does not imply global desirability. Evidence on infinite agentic loops makes termination a software reliability issue rather than a purely linguistic one [13].

### 2.2 Cost, budgets, and bounded computation

Classical bounded rationality and metareasoning treat computation itself as a decision resource [15,16]. Modern LLM work reaches the same question empirically. Token-budget-aware reasoning, CostBench, BAGEN, DeepPlanning, and test-time overthinking studies expose mismatches between capability, cost, and efficient allocation of computation [5-7,9-11]. AI Agents That Matter argues that evaluation should jointly consider accuracy and cost [4]; the 2026 agent-evaluation survey similarly identifies cost efficiency as a major gap [8]. GEC is complementary: it focuses on project-level scope, evidence, and stopping rather than token compression alone.

### 2.3 Verification and termination

Verification-aware planning encodes passing criteria into planning [17]. Progress-mirage work shows that self-evaluation can accept cycles that do not improve external world state [14], while Evidence-Carrying Termination makes terminal success contingent on typed support [18]. GEC extends this logic by treating evidence as state-versioned: evidence that was valid before a later change can become stale and must not remain a permanent completion certificate.

## 3 The Clinical Metaphor and Its Limits

### 3.1 Why "Parkinsonism" is used only as a phenomenological analogy

Parkinson disease (PD) bradykinesia is characterized by slowness, reduced movement amplitude, and the sequence effect [19]. The sequence effect describes progressive decrement in speed or amplitude of repetitive movement and has been studied experimentally in PD [20]. One phenotype of gait festination involves progressive shortening of step length with compensatory increase in cadence [21].

The analogy here is intentionally narrow. We do not claim that LLMs have basal-ganglia dysfunction, dopamine deficits, biological motor-control pathology, or any mechanism homologous to PD. The analogy concerns only a possible surface trajectory: large early task steps, lower-yield later actions, a plateau in verified utility, and persistence after the economic case for continuation weakens.

Figure 1 is therefore conceptual. Action granularity is not treated as a validated benchmark endpoint in LPB v0.2. It remains a motivating phenomenological signature for future live-model study, because cross-domain measurement of "action amplitude" is substantially less reliable than externally verified task utility and token cost.

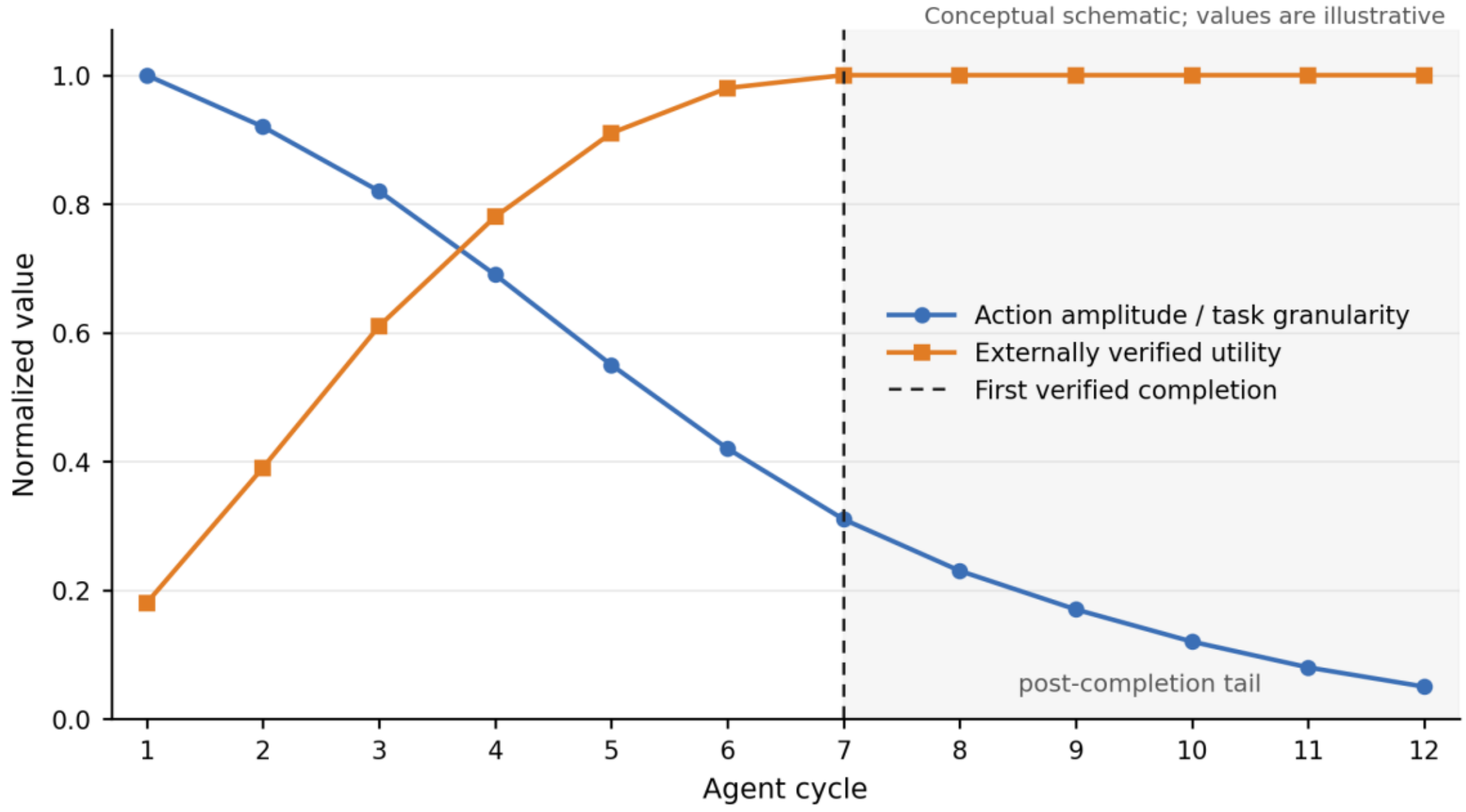

Figure 1. Conceptual trajectory motivating the "LLM Parkinsonism" metaphor. Action amplitude/task granularity is illustrative and is not treated as a validated LPB v0.2 endpoint. The empirically evaluated variables are governed scope, externally evaluated utility, token cost, complexity, evidence validity, and stopping.

### 3.2 Operational definition

We define LLM Parkinsonism as an agent-level trajectory exhibiting a substantial combination of four properties:

1. Goal drift: executed work loses a governed causal connection to unmet hard requirements, valid prerequisites, verification needs, risk-mitigation constraints, or explicitly authorized soft objectives.
2. Complexity accretion: the trajectory introduces additional states, components, dependencies, interfaces, or abstractions whose maintenance creates further work.

3. Termination failure: the system continues after the hard goal is externally verified, or continues when the best available governed continuation has non-positive marginal value.
4. Token-inefficient persistence: an increasing fraction of computation produces little or no externally verified task or process progress.

A single verbose answer does not establish this construct. It concerns multi-step, tool-using, project-level trajectories.

## 4 Why Autoregression Is Relevant but Insufficient

### 4.1 EOS is not PROJECT DONE

At generation time, an autoregressive model estimates

$$P(x_{t+1} \mid x_{1:t}).$$

An agent loop introduces a higher-order transition:

$$S_t \rightarrow A_t \rightarrow S_{t+1} \rightarrow A_{t+1}.$$

The question "should this response end?" is not equivalent to "should the project stop?" An orchestration layer can immediately request another step after clean response termination. A prompt such as "What should we do next?" pragmatically presupposes that a next step exists. A helpful model can satisfy that premise by finding some plausible improvement. In a sufficiently complex system, finding *a possible next action* is much easier than establishing that *a worthwhile next action* exists.

### 4.2 Self-conditioning can promote suggestions into requirements

Agents repeatedly condition on their own history. An optional suggestion can therefore enter a requirement-inflation pathway:

$$\textit{optional suggestion} \rightarrow \textit{history} \rightarrow \textit{assumed project fact} \rightarrow \textit{dependency} \rightarrow \textit{test} \rightarrow \textit{hard gate.}$$

A statement such as "we could add site-level alert aggregation" is initially optional. Several turns later, the same system may read the suggestion as a committed feature and generate tests, monitoring, and repair work for it. We call this self-generated requirement promotion.

### 4.3 Helpfulness can create action bias

Instruction-following correctly rewards useful responsiveness, but in an open-ended loop the bias toward a concrete next action can conflict with sufficiency. "Nothing further is required" is a legitimate project outcome, yet it may appear less action-oriented than another recommendation. This is a systems-level interaction between helpfulness and repeated next-step solicitation, not a claim that one particular training algorithm deterministically causes continuation.

### 4.4 Local coherence can defeat global objective review

Once a trajectory enters a subproblem such as "debug site correlation," successive steps can be locally coherent: adjust a timing window, revise suppression logic, add a synthetic test, repair the test. What is missing is a forced return to a higher abstraction level: why is site correlation part of the governed objective at all?

### 4.5 Self-evaluation can confuse activity with progress

A loop in which the same model proposes, executes, and judges a change creates a circular evaluator. Empirical work on progress mirages shows that self-evaluation can accept cycles with zero or negative measured change in external world state [14]. Thus:

$$\textit{Activity} \neq \textit{Progress.}$$

A new commit, test, paragraph, abstraction, or tool call is observable activity. Progress requires an independently grounded change in completion state, validated prerequisite state, risk state, or uncertainty.

### 4.6 Missing marginal-value computation

Classical metareasoning asks whether additional computation is worth its opportunity cost [16], consistent with bounded-rationality arguments [15]. Suppose an action has expected verified utility gain E[Δ U], token cost $C_T$, complexity cost $C_K$, operational risk $C_R$, and verification cost $C_V$. GEC uses an action-level score

$$V(a \mid s) = E[\Delta U(a \mid s)] - \lambda_T C_T(a) - \lambda_K C_K(a) - \lambda_R C_R(a) - \lambda_V C_V(a).$$

The crucial correction in v0.2 is logical: $V(a \mid s) \leq \tau$ means reject this candidate; it does not mean stop the project. Project-level economic stopping requires a state-level judgment over a candidate set or a justified continuation upper bound.

## 5 Formalization and Metrics

### 5.1 Governed Goal Contract

We represent a project as

$$G = (g, R_H, R_S, N, C, M).$$

where:

- g is the primary goal;
- $R_H$ is the finite set of hard requirements;
- $R_S$ is the set of explicitly authorized soft objectives;
- N is the set of explicit non-goals;
- C is the set of represented safety and feasibility constraints; LPB v0.2 does not yet exercise a dedicated general-purpose constraint predicate engine; and
- M is a contract-amendment policy.

The distinction between $R_H$ and $R_S$ resolves an ambiguity in the earlier formulation. A required criterion cannot be abandoned merely because it is inconvenient. If hard requirements remain unmet and no feasible path exists, the appropriate terminal state is BLOCKED, not GOOD ENOUGH. Economic stopping applies after hard completion when optional/soft continuation is not worth its cost.

Each hard criterion $r_i$ has weight $w_i > 0$. Externally evaluated hard-goal utility is

$$U_H(s) = \frac{\sum_i w_i v_i(s)}{\sum_i w_i}.$$

where $v_i(s)$ is determined by the external benchmark oracle or live verifier, not by the executor's narrative.

A proposal from the LLM is not a project requirement:

$$\textit{LLM proposal} \neq \textit{Goal Contract}.$$

Changes to the contract are explicit amendment events authorized outside the executor. The reference implementation increments a contract revision and rejects anonymous/self-authorized amendments.

### 5.2 Governed action linkage

The earlier rule "Which criterion does this action close?" was too strict for complex workflows. Necessary work can be enabling rather than immediately criterion-closing. GEC v0.2 therefore uses an independent scope assessment

$$L(a, r_i) \in \{DIRECT, PREREQUISITE, VERIFICATION, RISK_MITIGATION, SOFT, NONE, FORBIDDEN\}.$$

A prerequisite action may legitimately have $\Delta U_H = 0$ at the current step while still being causally necessary for later closure. NONE and FORBIDDEN actions are unscoped unless the Goal Contract is formally amended.

The generator's self-declared link is not authoritative. The architecture inserts an independent scope adjudicator between proposal generation and project governance. In a live system this could be a deterministic dependency graph, rule engine, separate model, human approval process, or ensemble; the key property is separation of authority.

### 5.3 Evidence validity and invalidation

For criterion $r_i$, GEC stores typed evidence

$$E_i = (status, verifier, state\ version, confidence, dependencies).$$

Completion requires evidence that exists, passes, exceeds a confidence threshold, and remains valid. Later changes can invalidate earlier evidence. This prevents a test that passed at state version k from remaining a permanent certificate after the relevant code or world state has changed at version k+m.

### 5.4 Primary outcomes and restricted completion cost

The two primary outcomes are:

1. Hard-goal success: all $R_H$ criteria are externally satisfied at terminal evaluation.
2. Restricted mean tokens to completion ($RMTTC_B$): a budget-restricted completion-cost endpoint defined over all episodes, where an episode that does not complete by ceiling B contributes B rather than being dropped from the analysis.

$$RMTTC_B = \frac{1}{n} \sum_{e=1}^{n} min(T_{c,e}, B), \ \ with\ T_{c,e} = \infty\ if\ episode\ e\ does\ not\ complete\ by\ B.$$

Conditional tokens to first completion among successful trajectories are retained as a secondary descriptive measure, because conditioning on success can induce selection bias when completion rates differ across policies.

Together, hard-goal success and $RMTTC_B$ compare both whether governance reaches the goal and how much restricted computation is required without discarding failed trajectories.

### 5.5 Token Efficiency

We retain

$$TE = \frac{1000\ U_H}{T_{total}}.$$

TE is useful for cross-budget comparisons, but it is secondary. When $U_H = 1$ for all compared trajectories, TE is simply $1000/T_{total}$ and therefore is not independent evidence beyond token cost.

### 5.6 Direct and process useful-token ratios

A strict direct useful-token ratio credits tokens only when a cycle closes previously unmet hard-goal utility:

$$UTR_{direct} = \frac{T_{direct}}{T_{total}}.$$

A broader process useful-token ratio also credits validated prerequisites, new valid verification evidence, and independently grounded process progress:

$$UTR_{process} = \frac{T_{process}}{T_{total}}.$$

This separation prevents a legitimate multi-step prerequisite chain from being mislabeled as pure waste merely because intermediate actions do not immediately change $U_H$.

### 5.7 Termination Overrun Ratio

Let $T_c$ be cumulative tokens at the first externally complete hard-goal state. Then

$$TOR = \frac{T_{total} - T_c}{T_{total}}.$$

for completed trajectories. TOR measures the post-completion tail.

### 5.8 Pre-completion Goal Drift Rate

To avoid double-counting post-completion persistence, goal drift is measured only before first completion:

$$GDR_{pre} = N(unscoped\ executed\ actions\ before\ completion) / N(all\ executed\ actions\ before\ completion).$$

TOR and GDRpre therefore address different questions: whether the agent drifts before finishing, and whether it fails to stop after finishing.

### 5.9 Gross and net complexity

Net complexity can hide thrashing. An agent that adds +10 complexity units and later deletes -10 has net zero complexity but has still paid the cost of an unnecessary excursion. We therefore report

$$K_{gross}{}^\wedge + = \sum_t max(\Delta K_t, 0).$$

and

$$K_{net} = \sum_t \Delta K_t.$$

Normalized gross and net indices divide these quantities by the required baseline complexity $K_R$. LPB v0.2 currently produces almost identical gross and net values because it does not yet model systematic delete-after-add cycles; therefore the proposed deletion-first rule is treated as an architectural principle, not as an experimentally established benefit in this benchmark.

### 5.10 Exploratory Executive Persistence Index

For descriptive visualization only, the implementation reports an Executive Persistence Index (EPI) combining nonproductive process tokens, TOR, $GDR_{pre}$, and gross complexity. The previous name "LLM Parkinsonism Index" is dropped to avoid implying clinical validation and to reduce double interpretation of a metaphorical label. EPI remains exploratory and is not a primary endpoint.

## 6 Hypotheses

The revised framework yields six hypotheses:

- H1 - Tail persistence: uncontrolled local loops will exhibit non-zero TOR after first verified completion.
- H2 - Pre-completion drift: uncontrolled loops will execute a non-zero fraction of actions with no governed link before completion; GEC's independent scope authority will reduce this fraction.
- H3 - Budget insufficiency: a token ceiling alone will not reliably select high-value work under a constrained budget.
- H4 - Governance beyond candidate search: a candidate-set local control should isolate the benefit of best-of-three access; GEC should preserve its hard-goal success while reducing restricted completion cost, token use, drift, and complexity relative to that control.
- H5 - Component specificity: removing independent scope authority or the marginal-value gate will degrade distinct trajectory properties rather than all components contributing identically.
- H6 - Robustness: moderate noise in scope linkage and external verification will degrade performance gracefully rather than catastrophically.

The stronger claim that unscoped-action probability increases monotonically as the number of unmet criteria approaches zero is not directly established by LPB v0.2 and is reserved for live trajectory analysis.

## 7 Global Executive Control v0.2

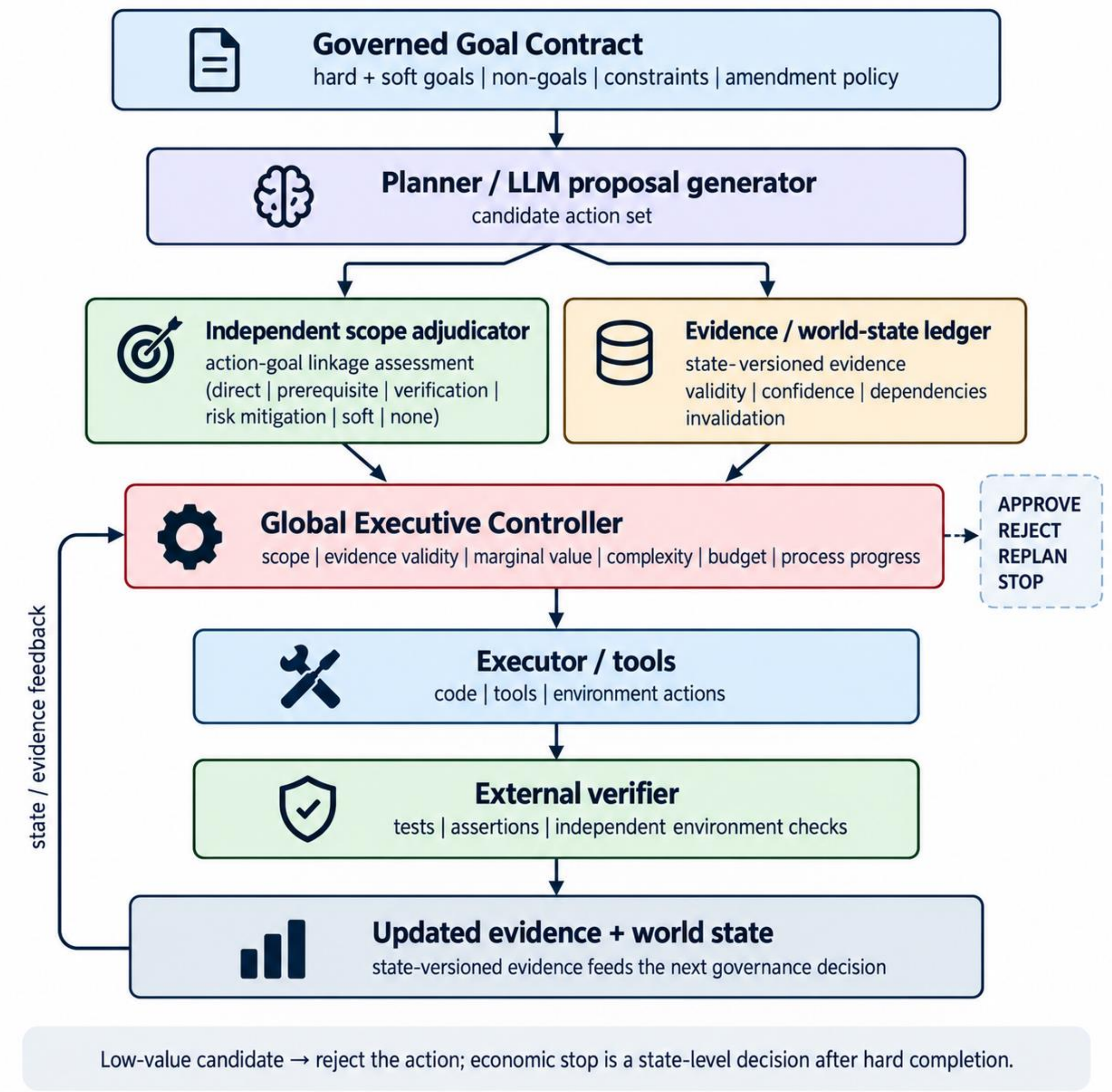

Figure 2. Global Executive Control v0.2. Proposal generation is separated from project-level authority through a governed Goal Contract, independent scope adjudication, state-versioned evidence, candidate-set value selection, and explicit terminal semantics.

### 7.1 Candidate-set governance

GEC v0.2 evaluates a set of candidate actions rather than treating the first proposal as destiny. This supports two important distinctions. First, a bad action can be rejected while another candidate remains viable. Second, economic stopping can be a state-level conclusion after evaluating available continuation options.

### 7.2 Independent scope gate

Each candidate receives a separate ScopeAssessment. Explicit non-goals produce FORBIDDEN; absent causal support produces NONE; direct, prerequisite, verification, risk-mitigation, and soft links can be eligible if confidence exceeds threshold. The scope gate therefore no longer trusts target_criterion supplied by the same LLM that proposed the action.

### 7.3 Contract-aligned expected utility

Expected hard utility is derived from criterion weight and candidate success probability rather than an arbitrary fixed utility constant. For a direct action linked to hard criterion $r_i$,

$$E[\Delta U_H] = p_{success}(a) \frac{w_i}{\sum_j w_j}.$$

Prerequisite and risk-mitigation links receive discounted continuation value because they are enabling rather than immediate completion events. Soft-objective value is separately weighted by β and cannot silently become hard utility.

### 7.4 Action-level rejection versus state-level economic stopping

For each candidate, GEC computes V(a|s). If V(a|s) ≤ τ, the action is rejected and the controller may choose another candidate or replan:

$$V(a \mid s) \leq \tau \Rightarrow REJECT(a), not\ STOP_PROJECT.$$

Economic stopping is allowed only when hard requirements are complete and no candidate in the governed continuation set has positive net value:

$$STOP_ECONOMIC\ iff\ R_H\ complete\ and\ max_a V(a \mid s) \leq \tau.$$

If hard requirements remain unmet and no feasible governed path exists, the correct state is STOP_BLOCKED. In the reference prototype, this terminal label requires a governed feasibility assessment supplied by the planner, environment, or an external feasibility oracle; GEC does not claim to solve feasibility inference internally. If the token budget is exhausted, the state is STOP_BUDGET.

### 7.5 Process-progress circuit breaker

The no-progress breaker now tracks more than binary criterion closure. It resets on hard-goal progress, validated prerequisite completion, or new valid external evidence. If no such progress occurs for k cycles, local repair is suspended and global replanning is requested. The prototype uses k=3.

Importantly, ablation results below show that this breaker does not improve efficiency in the current matched exogenous benchmark, because replanning cannot alter the future candidate stream by construction. We retain the mechanism because it targets a real live-agent failure mode, but we do not claim synthetic evidence for its independent benefit.

### 7.6 Evidence-carrying termination

STOP_SUCCESS requires valid evidence for every hard criterion. Evidence can be invalidated by later state changes. This creates a stronger done boundary than a boolean flag and protects against stale success certificates.

### 7.7 Governed amendments and non-goals

Non-goals are executable contract elements rather than documentation alone. A candidate matched to an explicit non-goal is rejected. If project priorities genuinely change, an externally authorized ContractAmendment produces a new revision. The executor may propose an amendment but cannot authorize it itself.

### 7.8 Deletion-first search

When a problem is caused by self-created structure, GEC conceptually prioritizes:

$$DELETE \rightarrow SIMPLIFY \rightarrow FIX \rightarrow ADD.$$

This principle is motivated by complexity control but is not quantitatively isolated in LPB v0.2; future benchmarks should include explicit add-delete-thrash trajectories.

## 8 Methods

### 8.1 Implementation

GEC v0.2 is implemented as a dependency-free Python research prototype. Core objects include GoalContract, ContractAmendment, AcceptanceCriterion, ScopeAssessment, ActionProposal, Evidence, TaskState, GlobalExecutiveController, metrics, and the LPB simulator. Thirteen unit tests cover evidence-dependent completion, non-goal rejection, prerequisite acceptance, candidate-level rejection, state-level economic stopping, beneficial soft work, governed contract amendment, no-progress replanning, separated drift/tail metrics, matched candidate streams, common budget ceilings, and simulation behavior.

### 8.2 LPB v0.2 task families

The benchmark contains six synthetic task families: two-node disaster-recovery watchdog, API release, database migration, CI pipeline, backup/restore verification, and reproducible research pipeline. Each scenario specifies three or four hard criteria, explicit non-goals, action-token distributions, prerequisite structure for selected criteria, and required baseline complexity.

LPB v0.2 is a mechanism-isolation testbed, not a prevalence estimator for real models.

### 8.3 Matched exogenous candidate sets

The principal design change from LPB v0.1 is policy-independent candidate generation. For each episode, the simulator pre-generates candidate sets and execution random draws from a common seed. Baseline, budget-only, and GEC receive the same candidate opportunities and stochastic outcomes. Governance can decide which candidate to execute or reject, but it cannot force the proposal generator to emit a required action on the next cycle.

Each cycle contains three candidates. The baseline local loop selects the first candidate. The budget-only policy preserves proposal order but skips a candidate only when it cannot fit in the remaining hard budget. A candidate-set local control sees the same three candidates and selects using only the generator's own declared action-to-goal links (preferring a declared link to an unmet governed criterion, with lower token cost as a tie-breaker), without independent scope adjudication, evidence-aware value gating, or complexity/risk penalties. GEC independently adjudicates scope, computes contract-aligned expected value, and selects the approved candidate with the highest net value. The candidate-set local control isolates the benefit of access to multiple candidates from the additional benefit of independent governance.

### 8.4 Candidate distribution

The default stream contains 28% optional proposals. Some optional proposals deliberately make misleading self-declared links to hard criteria, allowing the scope-authority ablation to test whether self-report can bypass governance. Among governed candidates, the stream includes direct requirement actions, prerequisite actions, verification actions, and a 10% branch of in-scope but deliberately expensive/low-yield verification candidates to make the marginal-value gate testable. Hard actions succeed with scenario probability 0.90 unless a required prerequisite has not been validated.

Each cycle consumes 120 planning tokens plus the selected action cost. Optional work can create maintenance debt and can stale prior evidence at low probability. After controller-perceived completion, the baseline local loop recognizes completion with probability 0.12 per cycle; GEC uses deterministic valid-evidence termination.

### 8.5 Governance conditions

Four main conditions are compared:

1. Baseline local loop: selects the first candidate, executes without global scope/value governance, uses probabilistic project-level stopping, and obeys the common hard token ceiling.
2. Budget-only: same local ordering with an explicit remaining-budget feasibility gate but no independent scope, evidence, or value governance.
3. Candidate-set local control: sees the same three-candidate set, selects using generator-declared links and a local tie-break rule, but has no independent scope authority, no evidence-aware marginal-value gate, and no complexity/risk governance.
4. GEC v0.2: independent scope adjudication, valid-evidence completion, contract-aligned value, complexity/risk/token penalties, candidate-set selection, and process-progress monitoring.

All four policies obey the same 40,000-token hard ceiling in the main benchmark.

### 8.6 Main experiment

The main benchmark uses six scenarios, four policies, and 1,000 episodes per scenario-policy cell: 24,000 episodes. The hard token ceiling is 40,000 tokens, the candidate-set horizon is 40 cycles, and the fixed root seed is 20260911. Binary hard-goal success is summarized with Wilson 95% confidence intervals;

continuous Monte Carlo means are reported as mean ± 1.96 SE conditional on the fixed synthetic scenarios and parameters.

### 8.7 Budget sensitivity

Budget-only and GEC are compared at 3k, 4k, 5k, 6k, 8k, 12k, 20k, and 40k tokens, with 150 episodes per scenario-policy-budget cell: 14,400 episodes. Success proportions use Wilson 95% confidence intervals, while continuous metrics use mean ± 1.96 SE. We also report a normalized area under the success-versus-log-budget curve:

$$AUC_B = \frac{\int P(success \mid B)\, d \log B}{\log B_{max} - \log B_{min}}.$$

This summarizes budget efficiency across the full resource range rather than at one arbitrary threshold.

### 8.8 Component ablation

Four policies are compared using 300 episodes per scenario: full GEC, GEC without independent scope authority (generator self-report is trusted), GEC without the marginal-value gate, and GEC without the no-progress breaker: 7,200 episodes.

### 8.9 Scope/verifier-noise robustness

GEC is tested with matched synthetic scope false-positive/false-negative noise and verifier noise at nominal levels 0, 0.02, 0.05, and 0.10, using 150 episodes per scenario-level cell: 3,600 episodes. This experiment is deliberately simple; it tests graceful degradation, not calibration to real verifier error rates.

### 8.10 Governance-overhead sensitivity

The main synthetic accounting does not assume that independent scope adjudication and candidate scoring are literally free. We therefore repeat the 40,000-token GEC experiment while charging an additional governance overhead $C_G \in \{0, 100, 250, 500, 1{,}000\}$ synthetic tokens per governed cycle, in addition to the fixed 120 planning tokens and executed-action cost. Using 1,000 episodes per scenario per overhead level yields 30,000 episodes. These charges are model-equivalent stress-test costs rather than calibrated provider prices.

### 8.11 Beneficial soft-work probe

A deterministic probe tests whether GEC incorrectly rejects all post-hard-goal work. After hard criteria are validly complete, the candidate set contains a low-cost beneficial soft action and a high-cost low-value soft action. GEC should approve the former, reject the latter, and continue rather than declaring economic stop from the single bad candidate.

## 9 Results

### 9.1 Main matched-candidate benchmark

All four policies face the same exogenous three-candidate sets and the same 40k ceiling. Their trajectories differ substantially.

| Metric | Baseline local loop | Budget-only | Candidate-set local | GEC v0.2 |
|---|---|---|---|---|
| Hard-goal success | 67.42% [66.22, 68.59] | 67.85% [66.66, 69.02] | 96.53% [96.04, 96.97] | 96.57% [96.08, 97.00] |
| Mean total tokens | 32,058 ± 207 | 32,170 ± 209 | 19,782 ± 184 | 12,574 ± 146 |
| $RMTTC_{40k}$ | 29,024 ± 262 | 29,021 ± 262 | 16,136 ± 195 | 13,114 ± 184 |
| Conditional tokens to first completion* | 23,719 ± 262 | 23,819 ± 263 | 15,279 ± 163 | 12,158 ± 137 |
| Direct Useful Token Ratio | 0.1122 ± 0.0015 | 0.1122 ± 0.0015 | 0.1954 ± 0.0023 | 0.3238 ± 0.0036 |
| Process Useful Token Ratio | 0.3163 ± 0.0027 | 0.3166 ± 0.0027 | 0.3614 ± 0.0030 | 0.4425 ± 0.0043 |
| Token Efficiency | 0.0317 ± 0.0004 | 0.0317 ± 0.0004 | 0.0589 ± 0.0007 | 0.0963 ± 0.0011 |
| Termination Overrun Ratio | 0.1249 ± 0.0042 | 0.1256 ± 0.0042 | 0.1967 ± 0.0044 | 0.0000 |
| Pre-completion Goal Drift Rate | 0.2721 ± 0.0023 | 0.2718 ± 0.0023 | 0.1290 ± 0.0021 | 0.0000 |
| Gross Complexity Accretion Index | 1.6543 ± 0.0161 | 1.6604 ± 0.0161 | 0.1228 ± 0.0013 | 0.1228 ± 0.0013 |
| Exploratory EPI | 46.07 ± 0.16 | 46.07 ± 0.16 | 41.09 ± 0.23 | 23.53 ± 0.18 |

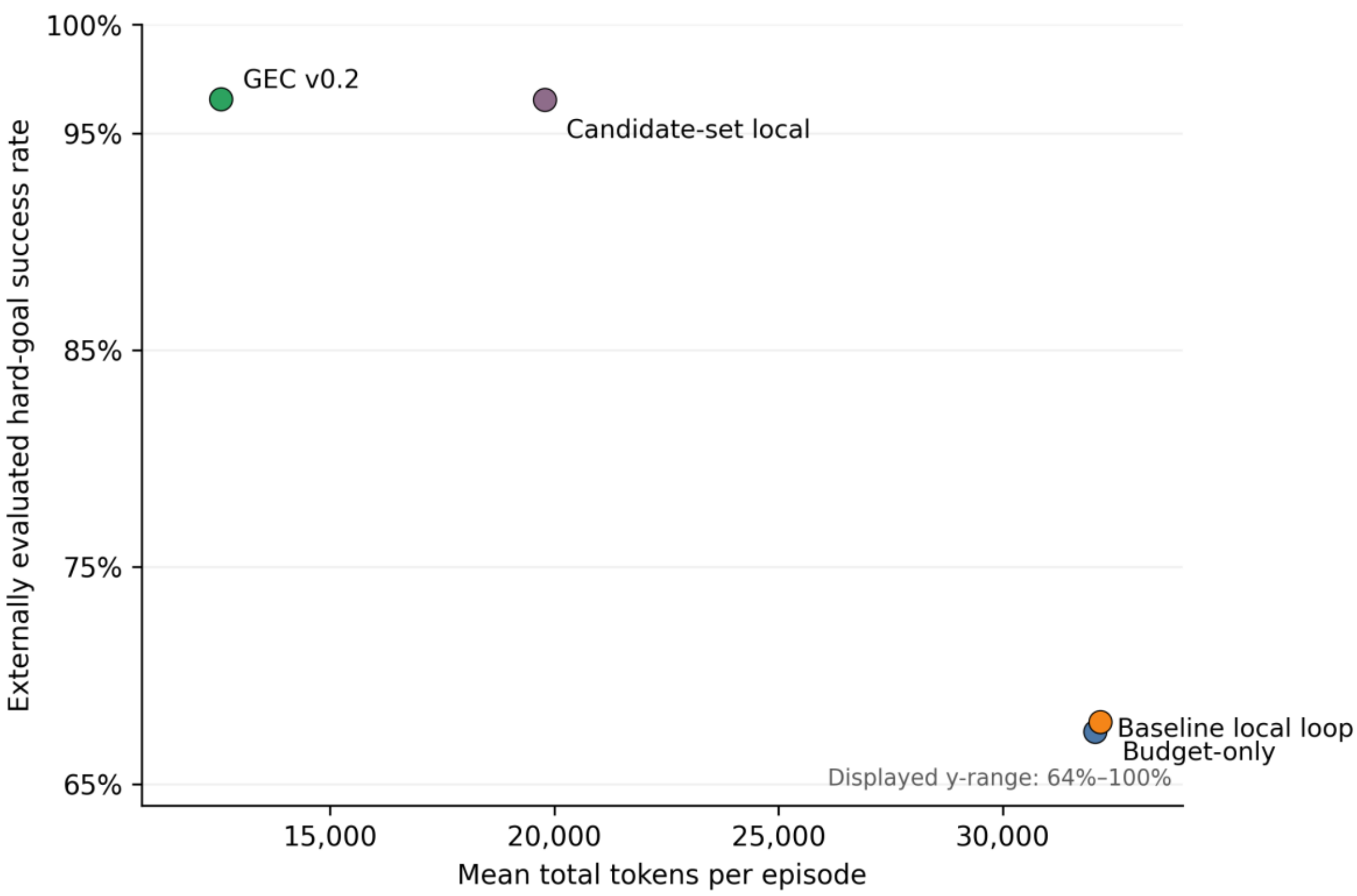

Figure 3. Matched-candidate benchmark operating points including the candidate-set local control. Candidate-set access recovers most of the hard-goal success gain, whereas GEC v0.2 shifts the operating point leftward at essentially the same success rate by reducing token use. The y-axis is truncated to 64–100% for visualization. Values are synthetic benchmark results, not measurements of a commercial LLM.

**Conditional tokens to first completion are calculated only among trajectories that complete and are therefore descriptive; $RMTTC_{40k}$ is the all-episode restricted primary completion-cost endpoint.*

Relative to the first-candidate baseline, GEC increased hard-goal success by 29.15 percentage points and reduced mean total token use by 60.8%. However, the candidate-set local control reached 96.53% success, essentially the same as GEC at 96.57%, showing that access to multiple candidate opportunities explains most of the success gain. Relative to the candidate-set local control, GEC reduced mean total tokens by 36.4%, reduced $RMTTC_{40k}$ by 18.7%, increased TE by 1.63×, reduced pre-completion GDR from 0.1290 to 0, and reduced gross complexity accretion from 0.8224 to 0.1228. The strongest causal evidence for GEC beyond best-of-three candidate access is therefore improved efficiency, scope discipline, and complexity control rather than additional hard-goal success.

Zero TOR under GEC is an enforcement consequence of deterministic hard-goal completion in this benchmark; zero pre-completion GDR is likewise substantially enforced by the default independent scope gate. These quantities validate controller invariants. The candidate-set local control is therefore essential: it shows that GEC does not obtain its main result merely from seeing three candidates, while also preventing us from attributing the entire success increase over the first-candidate baseline to governance.

### 9.2 Budget-sensitivity advantage

GEC increasingly outperformed budget-only control as budgets became large enough for governed completion pathways to fit:

| Budget | Budget-only success | GEC success | Budget-only TE | GEC TE |
|---|---|---|---|---|
| 3,000 | 0.00% | 0.00% | 0.0544 | 0.0769 |
| 4,000 | 0.00% | 0.78% | 0.0477 | 0.0785 |
| 5,000 | 0.11% | 3.22% | 0.0485 | 0.0782 |
| 6,000 | 0.44% | 7.11% | 0.0472 | 0.0806 |
| 8,000 | 1.67% | 22.44% | 0.0447 | 0.0834 |
| 12,000 | 6.89% | 54.11% | 0.0409 | 0.0897 |
| 20,000 | 26.11% | 87.00% | 0.0357 | 0.0945 |
| 40,000 | 68.67% | 96.33% | 0.0316 | 0.0956 |

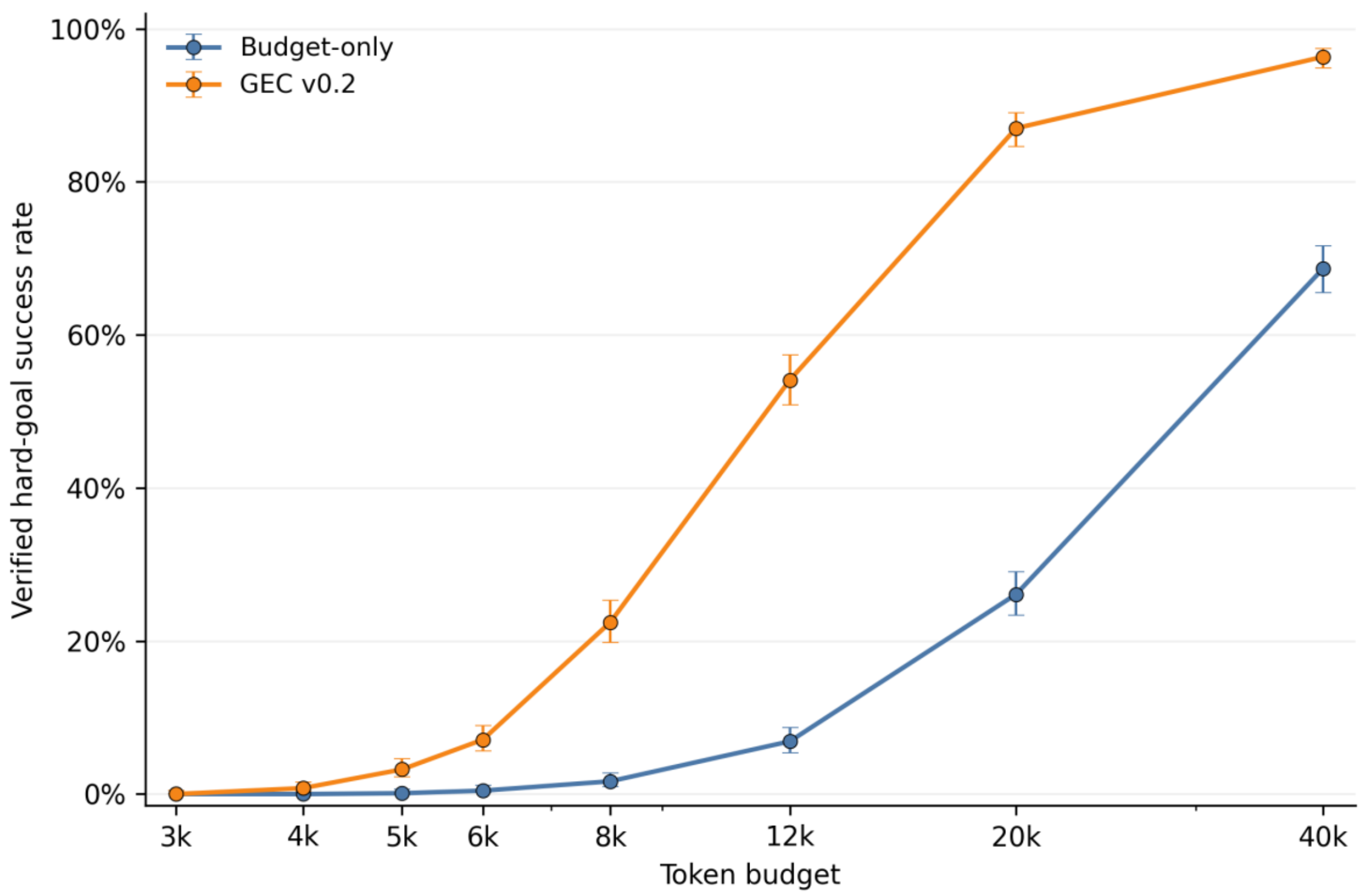

Figure 4. Hard-goal success under token ceilings. Error bars are exact Wilson 95% confidence intervals for 900 simulated episodes per policy-budget cell; the x-axis is logarithmic.

The normalized log-budget success AUC was 0.467 for GEC and 0.167 for budget-only control, a 2.79× ratio. The interpretation is not that GEC creates capability ex nihilo; rather, it allocates a fixed candidate opportunity stream toward actions with governed causal relevance and higher expected value.

### 9.3 Component ablations

| Policy | Success | Mean tokens | TE | GDR_pre | Gross CAI |
|---|---|---|---|---|---|
| Full GEC | 0.9611 | 12,733 | 0.0951 | 0.0000 | 0.1243 |
| No independent scope authority | 0.9522 | 14,036 | 0.0874 | 0.0716 | 0.3144 |
| No value gate | 0.9600 | 14,500 | 0.0862 | 0.0000 | 0.1263 |
| No no-progress breaker | 0.9611 | 12,318 | 0.0977 | 0.0000 | 0.1243 |

Removing independent scope authority produced measurable pre-completion drift and more than doubled gross complexity relative to full GEC. Removing the value gate preserved success but increased mean token use by about 1.77k tokens. By contrast, removing the no-progress breaker slightly improved synthetic efficiency because LPB v0.2 uses an exogenous candidate stream that replanning cannot change. This is an important negative result: the current benchmark does not establish an independent benefit of the no-progress breaker. Its value should be tested in live settings where replanning can alter subsequent proposals.

### 9.4 Robustness to imperfect governance signals

At injected noise levels from 0 to 0.10, GEC success remained between 96.33% and 96.00% in the robustness experiment, and TE remained approximately 0.095-0.096. The candidate-set design provides redundancy: a false negative on one candidate need not terminate the project if another valid candidate is available. These results should not be interpreted as real-world verifier calibration; they only show that the synthetic controller does not collapse under modest random errors.

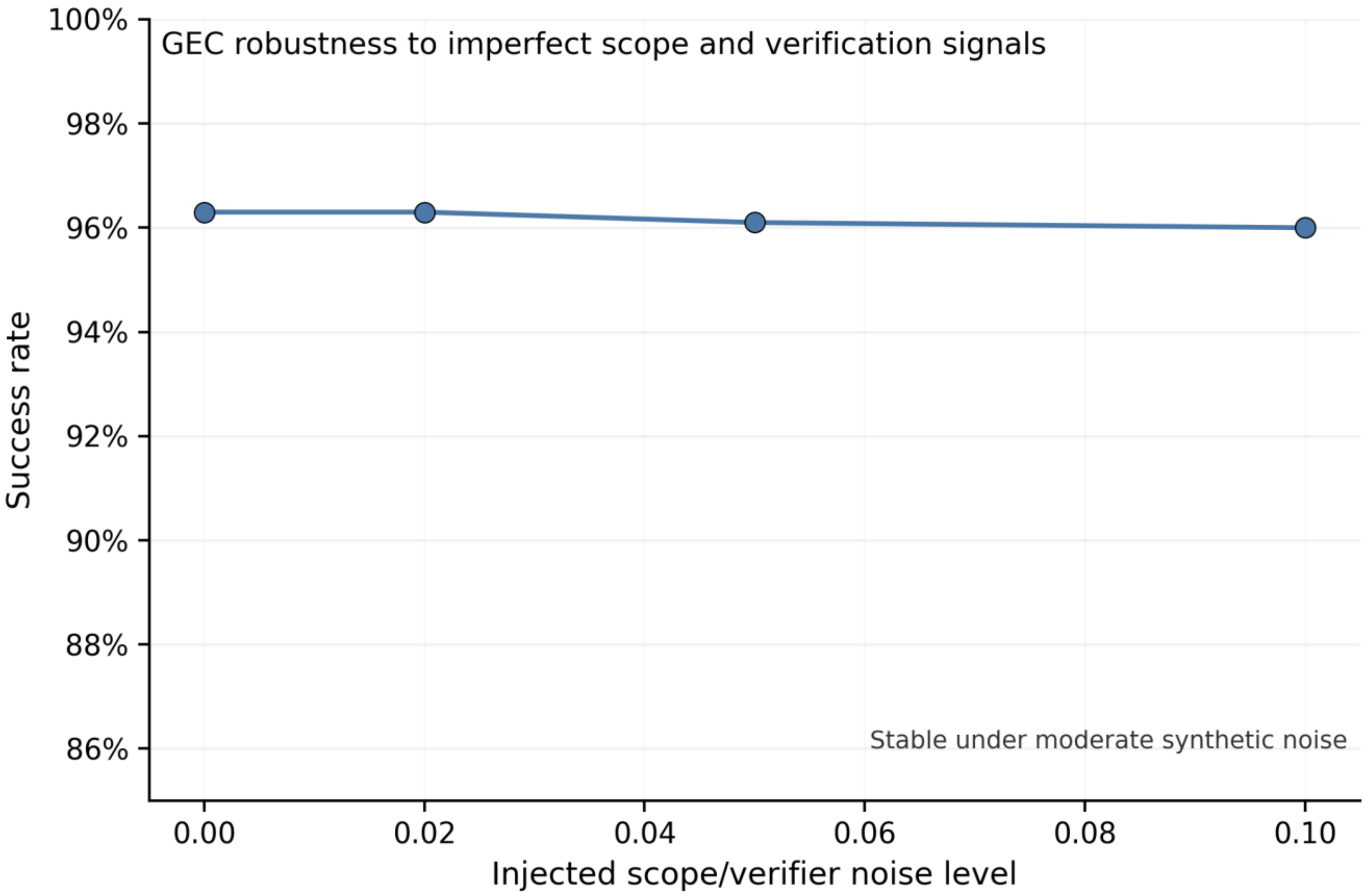

Figure 5. GEC v0.2 robustness under synthetic scope/verifier noise. The injected error model is illustrative and should not be interpreted as calibrated real-world verifier performance. The y-axis is truncated to approximately 85–100% for visualization.

### 9.5 Governance-overhead sensitivity

Charging nonzero synthetic governance cost reduces the apparent advantage, as expected, but the effect is gradual. GEC retained a clear success and restricted-completion-cost advantage over the first-candidate baseline through 500 additional governance tokens per cycle; at 1,000 tokens per cycle its success approached baseline and $RMTTC_{40k}$ became slightly worse than baseline. This sensitivity analysis bounds how much hidden adjudication cost the synthetic result can absorb without claiming that any one overhead level corresponds to a real provider.

| $C_G$ per cycle | Success (95% Wilson CI) | Mean total tokens | $RMTTC_{40k}$ |
|---|---|---|---|
| 0 | 96.57% [96.08, 97.00] | 12,574 ± 146 | 13,114 ± 184 |
| 100 | 96.55% [96.06, 96.98] | 14,842 ± 170 | 15,216 ± 194 |
| 250 | 95.93% [95.40, 96.40] | 18,188 ± 204 | 18,353 ± 213 |
| 500 | 89.88% [89.09, 90.62] | 23,251 ± 239 | 23,281 ± 240 |
| 1,000 | 67.08% [65.88, 68.26] | 30,320 ± 230 | 30,494 ± 234 |

At $C_G$ = 500 tokens per cycle, GEC still achieved 89.88% success and $RMTTC_{40k}$ = 23,281, compared with 67.42% and 29,024 for the first-candidate baseline. At $C_G$ = 1,000, the advantage largely disappeared (67.08% success; $RMTTC_{40k}$ = 30,494).

### 9.6 Beneficial soft work is not automatically suppressed

In the deterministic soft-objective probe, GEC approved the low-cost beneficial soft action (net value 0.2321), rejected the expensive low-yield soft action (net value -0.1605), and returned CONTINUE rather than STOP_ECONOMIC. This directly tests the corrected distinction between rejecting one candidate and stopping the project.

### 9.7 Watchdog case study

The watchdog contract contains four hard behaviors and explicitly lists site aggregation and individual-alert suppression as non-goals. A proposal for SITE DOWN aggregation may falsely claim that it supports an existing criterion. Under GEC v0.2 the generator's claim is not enough: the independent scope assessment marks the proposal as forbidden or unlinked, so the action is rejected. By contrast, a staging or synchronization step that is independently judged to be a prerequisite for a hard criterion can be admitted even if it does not close the criterion immediately.

## 10 Discussion

### 10.1 The deeper failure is executive, not merely linguistic

The failure mode is often described informally as verbosity, but long-horizon agents can make the more expensive error one level above language generation. The system can lack reliable control over scope, evidence, value, and termination. Autoregression makes continuation locally natural, but does not uniquely cause the phenomenon; a non-autoregressive planner could also over-plan. The more general weakness is concentration of authority in a system that is simultaneously asked to propose work, decide whether it is required, perform it, judge whether it improved the world, and decide whether to continue.

### 10.2 "Can do" is not "should do"

As capability expands, the action set grows. A stronger model can discover more possible refactors, tests, fallbacks, and monitoring strategies than a weaker model. Without governance, a larger action set can increase cost rather than utility. Option availability does not imply option desirability.

### 10.3 Prerequisites matter: scope is causal, not merely lexical

The earlier direct criterion-closing gate was too aggressive. Real engineering contains enabling work. A good executive controller must distinguish causally necessary intermediate action from self-invented scope. This is why GEC v0.2 uses typed linkage and process progress. The architecture should become more conservative only where the causal case is weak, not where utility is delayed by a legitimate multi-step chain.

### 10.4 Independent scope authority matters

If the same generator can label its own proposal "required," scope gating becomes circular. The ablation supports this architectural concern: trusting generator self-report reintroduced pre-completion drift and complexity. In real systems the independent linker need not be perfect, but it should be organizationally or computationally separate from the proposing actor.

The candidate-set local control adds an important causal qualification: simply seeing three alternatives raised success to nearly the GEC level. GEC therefore should not be credited with the full success increase over the first-candidate baseline. Its distinct contribution in LPB v0.2 is that independent scope and value governance achieves comparable success with substantially lower token cost, lower restricted completion cost, less drift, and less complexity.

### 10.5 Economic stopping is a state-level decision

A low-value proposal provides evidence against that proposal, not against the existence of every alternative. This distinction is central. GEC v0.2 does not infer STOP_ECONOMIC from one candidate. Economic stopping requires hard completion plus a governed candidate set or continuation upper bound with no positive net-value option. This change makes the decision rule compatible with ordinary search and replanning.

### 10.6 Evidence has a lifetime

External verification is not a one-time ceremony. If later actions modify the relevant world state, evidence can become stale. State-versioned evidence and invalidation move the architecture closer to real software assurance, where a previously passing test may need to be rerun after a dependency or interface changes.

### 10.7 Success and restricted completion cost should dominate TE

TE remains useful, especially across heterogeneous partial-success trajectories and budgets. But when final utilities are equal, it is an inverse-cost transformation. Conditional tokens to first completion can also be biased when policies have different success rates because failed trajectories are excluded. Therefore the principal live comparison should pair externally adjudicated hard-goal success with a restricted-mean completion-cost or survival-style endpoint over all trials; conditional completion time should remain secondary.

### 10.8 A budget is an accounting constraint; governance is a decision rule

A hard ceiling asks whether the system is allowed to spend more. It does not determine which available action deserves the next token. The budget-sensitivity curve and $AUC_B$ make this distinction explicit. A rational agent should often stop far before the maximum because the task is done, and should sometimes continue under a large remaining budget because an unmet hard requirement still has a feasible positive-value path.

### 10.9 Complexity deletion is a first-class operation, but remains under-tested

Generative systems are optimized to produce. Senior engineering often succeeds by removing unnecessary structure. Gross complexity accounting makes add-then-delete thrashing visible even when final net complexity returns to zero. LPB v0.2 implements the metric but does not yet include a dedicated delete-thrash task family, so deletion-first search remains a design hypothesis rather than a benchmark-supported causal claim.

### 10.10 From oracle governance to uncertainty-aware governance

A naive executive controller assumes that it knows the true scope, progress state, and evidence status. Real governance is uncertain. GEC v0.2 begins to model that uncertainty by separating generator claims from scope adjudication, attaching confidence to links and evidence, and testing noise. The deeper research direction is not "GEC knows what is correct," but GEC governs uncertainty about scope, progress, evidence, and stopping.

### 10.11 Training implications

The current prototype is inference-time governance, but the principles suggest training objectives of the form

$$R = R_{success} - \lambda_T T - \lambda_K K_{gross} - \lambda_D D_{drift} - \lambda_A A_{unnecessary} + \lambda_S R_{correct\,stop}.$$

Training data should include completed projects where the optimal next action is stop, partially complete projects where continue is correct, prerequisite chains where intermediate $\Delta$ U is zero but causal progress is real, soft-objective settings where selective continuation is rational, and blocked projects where escalation is correct. Whether these behaviors can be internalized robustly or deterministic external governance remains necessary for high-stakes systems is an open question.

## 11 Threats to Validity and Limitations

### 11.1 Synthetic construction

LPB v0.2 is synthetic. Optional-action probability, action success, token-cost distributions, prerequisite structure, evidence invalidation, and baseline stopping propensity are stress-test parameters, not fitted estimates of a named LLM. Numerical savings and success differences are demonstrations under this construction, not forecasts of commercial-agent behavior.

### 11.2 Controller invariants are partly tautological

Zero TOR after valid completion is a controller invariant; zero pre-completion GDR under perfect independent scope adjudication is also substantially enforced. These outcomes show that the implementation obeys its rules. The more informative result is whether those rules improve externally evaluated success and cost under matched action opportunities.

### 11.3 Candidate-set abstraction

LPB v0.2 improves causal comparison by matching candidate sets, and the candidate-set local control separates access to multiple proposals from independent governance. Real agents, however, often generate candidates adaptively from their evolving state. In addition, real independent scope adjudication and candidate scoring have inference cost. The governance-overhead sensitivity explicitly charges synthetic per-cycle costs, but those costs are not calibrated to model-equivalent provider usage. Live studies should record the full cost of the generator, scope adjudicator, verifier, and controller.

### 11.4 Scope and evidence remain simplified

The independent scope adjudicator is an oracle-like synthetic component with controlled noise. Real scope linkage may be semantically ambiguous, and evidence can be incomplete or adversarial. The robustness experiment tests random error, not systematic bias, correlated errors, or strategic self-justification.

### 11.5 Utility, prerequisites, constraints, and feasibility are simplified

Real projects contain partially observable goals, changing priorities, interacting constraints, and long prerequisite DAGs. GEC v0.2 supports governed amendments and prerequisite links, but LPB uses short synthetic chains and simplified criterion weights. The Goal Contract schema represents safety/feasibility constraints C, but LPB v0.2 does not yet exercise a dedicated general-purpose constraint gate. Likewise, STOP_BLOCKED consumes a governed feasibility assessment supplied by the environment/planner/oracle rather than internally proving that no feasible path exists.

### 11.6 No-progress breaker not supported by the current ablation

The no-progress mechanism is retained for live-agent testing, but the matched exogenous benchmark does not show an efficiency benefit; removing it modestly improved TE. It should therefore not be cited as a validated component effect from LPB v0.2.

### 11.7 Deletion-first search not isolated

Gross and net complexity metrics are implemented, but the benchmark does not yet contain systematic add-delete thrashing. A dedicated complexity-reversal benchmark is needed.

### 11.8 Action granularity is conceptual

The sequence-like reduction in action amplitude motivates the metaphor, but LPB v0.2 does not claim a validated cross-domain action-granularity metric. Live trajectory studies could develop one, but the current operational construct rests on scope, evidence, utility, cost, and stopping variables.

### 11.9 The medical metaphor can be misunderstood

"LLM Parkinsonism" is memorable but scientifically risky if interpreted literally. The paper repeatedly limits the analogy and does not propose biological homology. A neutral alternative is agentic executive-control persistence. Future work should test whether the metaphor aids conceptual clarity without creating avoidable clinical misunderstanding.

## 12 Prespecified Live-Model Validation

A credible next step is a paired experiment with real tool-using models. For each task instance, the same base model, temperature/reasoning setting, tool surface, Goal Contract, and acceptance criteria should be used under baseline, budget-only, and GEC governance. Exogenous task states and common randomness should be shared where technically feasible.

The primary endpoints should be:

1. externally adjudicated hard-goal success with a prespecified non-inferiority or superiority hypothesis; and
2. a restricted mean tokens-to-completion (or equivalent token-index survival/RMST analysis) over all paired trials up to a prespecified token ceiling, so non-completion is not silently excluded.

Secondary endpoints should include conditional tokens to first verified completion among successful trials, TE, TOR, $GDR_{pre}$, gross and net complexity, tool-call count, wall time, contract-amendment frequency, verifier invalidation events, and heavy-tail token percentiles. Exact input/output tokens should be recorded; if a provider exposes separate reasoning-token accounting, provider-billed total tokens and the available breakdown should both be reported. The full inference cost of any independent scope adjudicator or verifier must be included.

Scope adjudication should be independently audited. At minimum, a sample of candidate-link decisions should be double-coded by humans or by a blinded secondary evaluator. Completion should not be adjudicated solely by the same LLM that proposed and executed the work when deterministic or external checks are available [14,17,18].

The live study should explicitly include: multi-step prerequisite tasks; tasks with legitimately beneficial soft improvements; changing-goal tasks requiring authorized contract amendments; tasks where later changes invalidate earlier evidence; and tasks where the correct outcome is BLOCKED rather than endless retry. Model families should include at least one frontier reasoning model and one lower-cost model. Paired bootstrap or permutation inference is appropriate for cost metrics, and P90/P95 tails should be reported because runaway trajectories are likely skewed.

## 13 Conclusion

The weakness examined here is not lack of capability. It is the possibility of high capability under weak executive control. An LLM agent can be excellent at answering "What can I do next?" while remaining unreliable at answering "Should I do anything next?"

GEC v0.2 treats project-level governance as a separate computational layer. The model proposes; the Goal Contract defines hard and soft objectives; an independent linker adjudicates causal scope; the verifier measures; evidence expires when relevant state changes; the controller selects among candidates; and stopping is a state-level decision rather than a side effect of one bad proposal.

LPB v0.2 also shows why causal controls matter: candidate-set access alone explains most of the hard-goal success gain over a first-candidate loop, while GEC's independent governance primarily improves efficiency, scope discipline, and complexity control at comparable success.

The compact rules are:

*No governed causal link ⇒ no mandatory action.*

*Low value of one action ⇒ reject that action, not the project.*

*Hard requirements complete + no positive-value governed continuation ⇒ economic stop.*

*Hard requirements incomplete + no feasible path ⇒ blocked, not success.*

Reliable long-horizon intelligence should therefore be evaluated not only by what an agent can do, but also by whether it preserves the objective, distinguishes prerequisites from scope drift, resists unnecessary complexity, maintains valid evidence, allocates computation economically, recognizes sufficiency, and stops for the right reason.

## 14 Code and Reproducibility Materials

The GEC v0.2 reference implementation, LPB v0.2 matched-candidate simulator, unit tests, experiment scripts, machine-readable results, architecture documentation, and live-evaluation protocol are maintained in the project repository: [https://github.com/DongshengXiao/LLM-Parkinsonism-Solutions](https://github.com/DongshengXiao/LLM-Parkinsonism-Solutions)